# An Advanced Two-Stage Model with High Sensitivity and Generalizability for Prediction of Hip Fracture Risk Using Multiple Datasets

## AUTHORS

Shuo Sun[1], Meiling Zhou[2], Chen Zhao[3], Joyce H. Keyak[4], Nancy E. Lane[5], Jeffrey D. Deng[6], Kuan-Jui Su[7], Hui Shen[7], Hong-Wen Deng[7], Kui Zhang[1,*], Weihua Zhou[8,9*]

[1] Department of Mathematical Sciences, Michigan Technological University, Houghton, MI, USA

[2] Department of Statistics, Grand Valley State University, Allendale, MI, USA

[3] Department of Computer Science, Kennesaw State University, Marietta, GA, USA

[4] Department of Radiological Sciences, Department of Biomedical Engineering, and Department of Mechanical and Aerospace Engineering, University of California, Irvine, CA, USA

[5] Department of Internal Medicine and Division of Rheumatology, UC Davis Health, Sacramento, CA 95817

[6] Geisel School of Medicine, Dartmouth College, Hanover, NH, USA

[7] Division of Biomedical Informatics and Genomics, Tulane Center of Biomedical Informatics and Genomics, Deming Department of Medicine, Tulane University, New Orleans, LA 70112

[8] Department of Applied Computing, Michigan Technological University, Houghton, MI, USA

[9] Center for Biocomputing and Digital Health, Institute of Computing and Cybersystems, and Health Research Institute, Michigan Technological University, Houghton, MI, USA

* Co-corresponding authors.

# Abstract

Hip fractures represent a major source of disability, mortality, and healthcare burden in older adults, making early risk assessment essential for effective prevention. However, widely used tools such as the Dual-energy X-ray Absorptiometry (DXA) T-score and Fracture Risk Assessment Tool (FRAX) are limited in sensitivity and often miss individuals who will experience a fracture, particularly those without prior fracture history or with osteopenia. These limitations highlight the need for new predictive approaches that can incorporate both clinical characteristics and imaging features to improve early detection. In this study, we present a novel sequential two-stage model for predicting hip fracture risk, using data from the Osteoporotic Fractures in Men Study (MrOS), the Study of Osteoporotic Fractures (SOF), and the UK Biobank. In Stage 1 of the model (Screening Stage), clinical, demographic, lifestyle, cognitive, and functional factors are used to estimate the baseline risk of hip fracture. In Stage 2 (Imaging Stage) of the model, imaging features from DXA scans are incorporated to further improve the prediction. The model's performance was rigorously evaluated through both internal and external validations, including independent testing in the UK Biobank, demonstrating its adaptability across diverse populations. This stepwise approach improved sensitivity compared to traditional tools like T-score and FRAX, enabling earlier and more accurate detection of individuals at high risk for subsequent hip fracture. By prioritizing early detection, this sequential model reduced missed cases by offering a more cost-effective strategy that reserves imaging for those identified as possibly at high risk in the initial screening evaluation for subsequent personalized fracture risk assessment.

# 1. Introduction

Hip fracture is a major public health concern in aging populations, leading to substantial morbidity, mortality, and healthcare costs[1]. With the growing number of older adults worldwide[2], the burden of hip fractures is expected to increase dramatically. Early identification of individuals at high risk of hip fracture is critical for timely intervention, since preventive treatments and lifestyle modifications can significantly reduce hip fracture risk. However, existing clinical tools for identifying individuals at high risk of hip fracture often lack sensitivity, resulting in a considerable number of high-risk individuals with osteopenia being missed, which in turn limits opportunities for effective prevention. Addressing this gap is crucial to reducing avoidable hip fractures, improving quality of life in older adults, and alleviating the societal and economic burden of osteoporosis-related injuries[3,4].

At present, bone mineral density (BMD) measurements via DXA using the T-score and the Fracture Risk Assessment Tool (FRAX) are the most widely used approaches for osteoporosis and fracture risk assessment. While valuable, these methods have important limitations, particularly low sensitivity[5,6]. The T-score relies solely on DXA BMD, overlooking other relevant risk factors, while FRAX incorporates clinical factors but does not directly leverage imaging-derived features or functional assessments. Both approaches often miss individuals with osteopenia or without a prior fracture history, leading to underdiagnosis and undertreatment.

Recent machine learning and ensemble modelling approaches have been explored to improve identification of individuals at high risk of hip fracture, but many models consider either clinical variables or imaging, without integrating information from both domains in a sequential, clinically efficient manner. For example, Nicholas[7] et al. (2018) showed that fall history independently predicted fractures in the Osteoporotic Fractures in Men Study (MrOS). In the Study of Osteoporotic Fractures (SOF), Schousboe[8] (2013) reported there was a modest association between cognitive impairment and false-negative identification of hip fractures. Webster[9] (2023), using UK Biobank data, found vegetarians had higher hip fracture risk, partially due to lower body mass index (BMI). Although prior research often considered clinical factors, BMD stands out as a robust predictor of osteoporosis and subsequent fracture risk[10] (Hong et al., 2020). Shaik[11] (2024) combined clinical factors with imaging (BMD) by introducing a staged machine learning model that improved hip fracture prediction. Although Shaik's models represented an important improvement, they lacked external validation. Without testing on independent cohorts or datasets, the generalizability and robustness of the proposed method across diverse clinical populations and settings remain uncertain.

This study advances the two-stage model framework originally proposed by Shaik et al. (2024). Their work demonstrated that their two-stage model outperformed their first stage (Ensemble 1) with respect to AUC but performed slightly inferiorly to their second stage (Ensemble 2), where "Ensemble" refers to bootstrapping strategy. Building on this, we now introduce a new uncertainty quantification rule that determines whether an individual's hip fracture risk assessment is completed in our Stage 1(Screening Stage) or if the individual's assessment advances to assessment using our Stage 2 (Imaging Stage). Beyond confirming the AUC results observed previously, our study compares the accuracy, sensitivity, and specificity of our advanced two-stage model with those of our Stage 1 Model alone and those of our Stage 2 Model alone.

In addition to advancing this two-stage methodology for identifying patients at risk of hip fracture, in this study, we move beyond internal evaluation, in which the model is developed and then evaluated on the same data set, to the critical step of external validation in which the model is developed on one data set and then evaluated on a different, independent data set. By directly contrasting the results from internal and external validation, we aim to demonstrate that, even when evaluated on an independent dataset, the two-stage model maintains strong performance in determining whether an individual's hip fracture risk can be reliably assessed using only Stage 1, or whether the individual should be further evaluated with Stage 2 for a more comprehensive risk assessment.

## 2. Methods

We developed and validated a two-stage machine learning framework for hip fracture risk assessment, based on data from three large population-based cohorts: MrOS, SOF, and the UK Biobank. Participants were carefully harmonized based on overlapping variables across all three cohorts to ensure comparability. To evaluate the performance of the two-stage model, we conducted both internal and external tests. Internal tests involved training and testing on the same dataset (MrOS for males, SOF for females), while external tests were trained on MrOS or SOF and validated on male or female cohorts, respectively, from the UK Biobank[12]. Our advanced two-stage model included formal uncertainty quantification and an adaptive refinement mechanism that enabled dynamic updating of risk estimates as additional imaging data was provided. Sensitivity for identifying high-risk individuals with our two-stage model was compared with that of conventional tools, including the WHO T-score thresholds for diagnosis of osteoporosis and the Fracture Risk Assessment Tool (FRAX) algorithm[13], to evaluate potential utility for pre-emptive intervention in clinical practice.

### 2.1 Data Sources

Three well-characterized cohorts were used in this study. MrOS and SOF are multicentre, prospective cohorts of older adults in the United States, comprising 5,994 men and 9,704 women aged ⩾65 years, respectively. Participants were community-dwelling and enrolled from geographically diverse U.S. regions. As described in the data sources[14-18], for MrOS, BMD was measured using the Hologic 4500, while SOF included both Hologic 1000 and Hologic 4500 measurements. For SOF, the conversion from Hologic 1000 to the equivalent Hologic 4500 scale was conducted by the original data providers; we refer to these BMD values as "Harmonized" to the Hologic 4500. To evaluate generalizability of our proposed model, we conducted an external validation using the UK Biobank (application ID: 61915), a large population-based resource with genetic and health data from approximately 500,000 individuals.[12,17] In the UK Biobank dataset, DXA scans were acquired using GE-Lunar iDXA machines with routine calibration and quality control[12,17].

### 2.1.1 Harmonized BMD and T-value BMD

To enhance reproducibility, we incorporated both harmonized BMD data, which were taken directly from the original data sources, and a T-value which was calculated from the BMD data across all analyses (T-value BMD). Although a T-score would compare BMD with that of healthy young adults of the same sex, the T-value here was compared to a healthy adult of the same sex in this cohort. Subjects were assigned to subgroups based on sex (male or female, as assigned at birth), age (65–75 and 76–96 years for MrOS and SOF[14-18]; 54–65, 65–70 years for UKBiobank), and race (MrOS and SOF race groups listed in Table 1, and all UK Biobank participants selected based on variables of interest were of White ancestry.).

In the internal test, we aimed to maintain consistency between MrOS and SOF, so both datasets used harmonized Hologic 4500 measurements or both datasets used T-values. In the external test, to ensure consistency across MrOS, SOF, and the UK Biobank, we computed T-values using the original data from each source—MrOS (Hologic 4500), SOF (Hologic 1000), and UK Biobank (GE-Lunar iDXA). Using the original measurements for T-value computation is essential to maintain consistency across datasets.

Using this T-value BMD instead of the T-score was necessary because MrOS and SOF did not include young individuals to serve as comparable reference data for computing a conventional T-score, we standardized each BMD value with respect to an age-, race-, and sex-specific subgroup of individuals without hip fractures to obtain the T-value BMD, which was calculated by subtracting the mean and dividing by the standard deviation of the corresponding subgroup without hip fracture,:

$$T-value = \frac{\text{BMD}_{\text{subject}} - \text{Mean BMD of subjects without hip fracture}}{\text{SD of subjects without hip fracture}}$$

We evaluated the harmonized BMD and T-value in the MrOS and SOF datasets (internal evaluations), and also calculated T-value for the UK Biobank (external evaluations). The following BMD variables were evaluated from MrOS: Total Hip BMD (B1THD), Total Spine BMD (B1TLD), and Femoral Neck BMD (B1FND), all from the Hologic 4500, From SOF we evaluated Total Hip BMD (HTOTBMD), Lumbar Spine BMD (STOTBMD), Femoral Neck BMD (NBMD), all from the Hologic 1000. For the UK Biobank, T-values were similarly computed for femoral neck, total hip, and spine BMD and stratified by age and sex. We used right hip measurements for males[16], and left hip measurements for females[17] to ensure consistency with the MrOS and SOF protocols which used right hip and left hip measurements, respectively.

## 2.3 Internal Evaluation for MrOS and SOF

### 2.3.1 Data Preprocessing & Feature Selection

To ensure high data quality, we extracted all the harmonized variables from the MrOS and SOF dataset and removed redundant features. Observations with missing hip fracture outcomes were excluded, and variables with more than 20% missing values were removed from the analysis. For the remaining data, given that the missing data rate was approximately 12.36% in the MrOS dataset and 11.23% in the SOF dataset, we opted to address the missing values by removing the corresponding observations. Numerical variables were normalized using a spatial sign transformation. Highly correlated continuous features (correlation > 0.83) were removed from the analysis to reduce multicollinearity. Univariate feature selection via near zero variance filtering was also employed. We also employed Recursive Feature Elimination (RFE)[20] to identify the most predictive variables in a multivariate context. Feature selection was conducted independently for each of the two stages. Over 100 iterations were used to calculate the average rank of each variable.

### 2.3.2 Evaluation of Stage 1 Model and Stage 2 Model

In Stage 1, which we call the Screening Stage, we implemented a logistic regression model based on clinical, demographic, wrist and spinal fracture records, and cognitive and functional variables to establish a baseline risk estimate for hip fracture. In Stage 2, the Imaging Stage, we incorporated quantitative imaging features extracted from hip dual-energy X-ray absorptiometry (DXA) scans into the model to refine risk estimates for hip fracture at the individual level. At each stage, we selected the most relevant predictor variables based on the feature selection steps in section 2.3.1, with the outcome variable, hip fracture, a binary variable for the logistic regression. To rigorously assess the performance of the logistic regression model, we employed a 100-run bootstrap sampling strategy combined with cross-validations. For each of the 100 bootstrapped samples, we generated a unique dataset by random resampling and then partitioned it into five folds for cross-validation in training and testing. This procedure constituted our "ensemble": each ensemble model was trained on a resampled subset of the data using stratified cross-validation, thereby promoting robustness and capturing variability within the dataset. The performance of each ensemble was then compared with that from a more comprehensive, advanced two-stage model in section 2.3.3.

Model performance was evaluated using AUC, accuracy, sensitivity, and specificity in each cross-validation fold, where the optimal classification threshold was determined using Youden's index, and a confusion matrix was used to compute the performance metrics. These metrics were averaged across five folds for each bootstrap sample. After all the 100 bootstrap runs, the final performance metrics were summarized by calculating the overall mean and standard deviation for each metric. This approach provided a robust and reliable assessment of the model's predictive performance and stability.

### 2.3.3 Evaluation of Advanced Two-Stage Model

To enhance predictive performance, a robust ensemble learning approach was employed based on the top variables identified in Section 2.3.2, integrating both clinical and DXA features, in an advanced two-stage model evaluation (Figure 1). The rationale behind this approach is that some cases evaluated using just the screening variable in Stage 1 may have low uncertainty and may be correctly classified without the use of Imaging variables from Stage 2, but cases with high uncertainty might be more effectively classified by incorporating DXA features. Therefore, uncertain cases were reassessed using Ensemble 2, which integrated DXA features alongside clinical variables. By selectively refining predictions for uncertain cases, this two-stage modeling approach sought to enhance overall classification accuracy while maintaining clinical efficiency.

To develop the Two-Stage model, we began with the original dataset and applied filtering based on the top variables identified in Stage 2 from section 2.3.2. Because some of these selected variables contained missing data, the dataset required further refinement by removing rows with missing values, resulting in the exclusion of approximately 11.01% of the SOF participants and 12.36% of the MrOS participants. The hip fracture was converted into a binary variable for logistic regression. The dataset was split into 80% training and 20% testing using stratified sampling, ensuring that the proportion of participants with and without hip fractures was preserved across both sets. This approach maintained outcome balance and allowed for more accurate and unbiased model evaluation.

Evaluation of the model used the following approach. First, Ensemble 1 was trained using only clinical features, employing a five-fold cross-validation strategy within the training data (80% of the entire dataset). In each round, one fold was used as the validation set while the other folds were used for training. Within each training fold, an ensemble of 50 bootstrapped logistic regression models was trained by repeatedly resampling the training data to create varied subsets. Each model generated probability predictions for the validation set, and the mean probability across all models was computed as the final prediction for that fold.

The optimal classification threshold (the best threshold) was determined using Youden's index by maximizing sensitivity and specificity. An ROC curve was constructed for each validation fold to assess model performance, and the Youden's index was calculated as the sum of sensitivity and specificity minus one (sensitivities + specificities - 1). The optimal classification threshold for each fold was determined by identifying the probability threshold that maximizes Youden's index. After cross-validation was completed, the best fold was selected as the one with the highest AUC value. The corresponding threshold from that fold was then chosen as the best decision threshold, ensuring that the model maximizes the trade-off between sensitivity and specificity.

The final ensemble model from Stage 1 was evaluated on the test set, for each participant in the test set, a z-score was calculated based on the predicted probability from Ensemble 1 that used only clinical variables. This uncertainty z-score is defined as:

$$uncertainty\ z-score = \left| \frac{mean\ of\ predicted\ probabilities\ from\ 50\ models - best\ threshold}{standard\ deviation\ of\ predicted\ probabilities\ from\ 50\ models} \right|$$

This z- score measured how far the predicted probability deviated from the optimal classification threshold, relative to its variability across the 50 bootstrapped models. A higher z-score indicated greater confidence in the prediction, while a lower z-score signified higher uncertainty. To systematically identify uncertain cases, a threshold of z = 2 was applied, meaning any participant with a z-score below this threshold was considered uncertain.  This threshold was chosen to capture cases where the predicted probability was within two standard deviations of the optimal threshold, a common statistical convention indicating moderate to high uncertainty. The total number of uncertain cases was then computed accordingly. Cases classified with high uncertainty were identified and re-evaluated using Ensemble 2, which incorporated both clinical and DXA features. The second ensemble applied the same training procedure as the first but incorporated additional DXA imaging features to enhance classification accuracy. The final predictions for the test set were obtained by merging results from both ensembles - predictions from Ensemble 1 were retained for confident classifications with low uncertainty, while uncertain cases were reassessed with Ensemble 2. The performance of the advanced two-stage model was assessed over 100 independent runs, measuring AUC, accuracy, sensitivity, and specificity. The final results were summarized by computing the mean and standard deviation for each metric.

## 2.4 External Evaluation for UK Biobank

### 2.4.1 Data Preprocessing

For external validation of the two-stage model, we used the top predictive variables from the second stage in the internal test to identify participants from the UK Biobank data set and evaluate our model, where DXA features were integrated with clinical variables. To ensure consistency across MrOS, SOF, and UK Biobank, we selected participants based on nine common clinical variables - grip strength, walking pace, history of upper arm/shoulder or wrist fracture, height, weight, age, smoking status, and difficulty walking long distances - along with three DXA-derived BMD measures: femoral neck, total spine, and total hip. Clinical and imaging variables were standardized by T-values. Data cleaning involved removing infrequent or ambiguous response categories (e.g., "Prefer not to answer") and recoding categorical variables into binary or numeric formats. Walking speed was categorized into tertiles, grip strength was averaged across trials, and IADL difficulty and smoking status were numerically encoded. These preprocessing steps ensured compatibility and comparability across all datasets for downstream modelling.

### 2.4.2 Evaluation of Advanced Two-Stage Model

We evaluated the two-stage model developed using the MrOS and SOF "training" datasets with the UK Biobank as an external test set. In the internal test datasets, follow-up time for hip fracture (Figure 3) was included as a covariate. However, since the UK Biobank dataset does not contain this variable, follow-up time was excluded when comparing model performance across all three datasets in the external validation. After aligning clinical and DXA features, a 100-run cross-validation was conducted, with performance evaluated on the UK Biobank using AUC, accuracy, sensitivity, specificity, and stage-wise accuracy to assess model generalizability and robustness.

# 3. Results

## 3.1 Cohort Characteristics and Key Predictive Features

Based on the data preprocessing and feature selection steps outlined above, the cohort characteristics and fracture prevalence of the MrOS and SOF datasets were summarized to inform the following analyses. Hip fracture prevalence was higher among women (17.9%, 491 out of 3239 participants) than men (3.7%, 134 out of 3764 participants), despite a larger male sample size. Racial differences (Table 1) were observed: in MrOS, the highest fracture rate occurred in the "Other" group (6.1%), while in SOF, Asian (20.0%) and Hispanic (33.3%) participants had elevated fracture rates, though sample sizes were small.

After applying RFE to identify the most predictive variables in a multivariate setting for MrOS and SOF datasets, the average variable ranks based on feature selection results were summarized in Figure 2 & 3. In Stage 1 (Figure 2), key predictors included age, BMI, physical performance, cognitive function, wrist and spinal fracture history. In Stage 2 (Figure 3), DXA-derived BMD measures (femoral neck, total spine, and total hip) were added as significant predictors of hip fracture. For UKBiobank, two external test datasets were constructed based on the selected variables: the first dataset with 1269 participants and 12 hip fracture subjects included total BMD values, and the second dataset with 625 individuals and 8 hip fracture subjects included side-specific total BMD values (Table 2).

## 3.2 Internal Tests: The Two-Stage Model Demonstrates Enhanced Performance

### 3.2.1 Superior overall performance of the two-stage model

For the MrOS male cohort, the two-stage model using harmonized BMD (Table 3) achieved markedly better performance than Ensemble 1 alone and Ensemble 2 alone. Specifically, the two-stage model reached an accuracy of 0.759, sensitivity of 0.712, and specificity of 0.761, outperforming Ensemble 1 (accuracy: 0.649, sensitivity: 0.340, specificity: 0.660) and Ensemble 2 (accuracy: 0.705, sensitivity: 0.279, specificity: 0.721). With T-value BMD (Table 4), the two-stage model further improved sensitivity to 0.717 while maintaining similar accuracy (0.759) and specificity (0.760), again surpassing both ensembles, which showed much lower sensitivities.

### 3.2.2 Balanced AUC performance explained by model design

The AUC of the two-stage model (harmonized BMD: 0.807; T-value BMD: 0.806) fell between that of Ensemble 1 (0.699) and Ensemble 2 (0.829). This outcome aligns with the model's sequential structure: predictions for individuals retained in Stage 1 rely on Ensemble 1, while those forwarded to Stage 2 depend on Ensemble 2. Thus, the two-stage model inherits strengths from both, while ensuring robustness across cohorts.

### 3.2.3 Consistent improvement in SOF female cohort

Similar trends were observed in the SOF female cohort. With harmonized BMD (Table 5), the two-stage model achieved higher accuracy (0.682), sensitivity (0.698), specificity (0.679), and AUC (0.761) than either Ensemble 1 (accuracy: 0.512, sensitivity: 0.483, specificity: 0.517, AUC: 0.605) or Ensemble 2 (accuracy: 0.593, sensitivity: 0.366, specificity: 0.634). Results were consistent when using T-value BMD (Table 6), where the two-stage model again delivered superior overall performance (accuracy: 0.697, sensitivity: 0.678, specificity: 0.701, AUC: 0.761).

### 3.2.4 Stage-wise accuracy illustrates efficiency of the approach

In our two-stage model, we applied the uncertainty rule ($z > 2$) to identify "certain" participants that enter Stage 1. These participants and their predicted outcomes are not always a perfect match, so we calculate accuracy accordingly. Because not all participants meet the $z > 2$ criterion, we determined the proportion of Stage 1 participants by dividing the number of certain participants ($z > 2$) by the total number of samples. Then we found that our stage-wise analysis demonstrated that the majority of participants could be classified early with high accuracy.

In MrOS, ~61.5–62.3% of individuals were classified in Stage 1 with an accuracy of ~0.846, while in SOF, ~74.5–75.2% were classified in Stage 1 with accuracy ranging from 0.727–0.743. The remaining participants, assessed in Stage 2, were handled with moderate but meaningful accuracy (MrOS: ~0.606–0.608; SOF: ~0.545–0.554). This two-stage design enables precise early classification for straightforward cases while allocating more complex cases to deeper evaluation, thereby balancing clinical accuracy with practical efficiency. Within this framework, even if stage 1 accuracy is higher than Stage 2 accuracy, it does not necessarily indicate that the Stage 1 model performs better. Our uncertainty rule is specifically designed to direct simpler (more certain) cases to Stage 1 and the more complex (less certain) ones to Stage 2—this is the main rationale for developing the two-stage approach. This suggests that the two-stage model employs an effective uncertainty rule for directing samples to Stage 2, demonstrating its potential value in clinical applications.

The internal validation results showed that the two-stage model delivered robust and consistent performance across both the harmonized BMD and T-value standardization approaches. Compared with the individual ensemble models, the two-stage framework achieved superior classification accuracy, sensitivity, and specificity in both the MrOS (male) and SOF (female) cohorts, underscoring its strong predictive value in internal validation.

## 3.3 External Tests: The Two-Stage Model Demonstrates Outstanding Performance

The advanced two-stage model results which utilized the T-value BMD data were evaluated using internal test datasets (MrOS and SOF) and externally validated with test data from the UK Biobank. The external test results from the UK Biobank with both 625 participants and 1269 participants demonstrated improved performance, particularly in terms of AUC and other key metrics.

### 3.3.1 Stage-Wise Accuracy Confirms Strong Cross-Cohort Performance

For the male model (Trained on MrOS), the UK Biobank test (Table 7) demonstrated substantially higher stage-wise accuracy in both stages. The average accuracy for Stage 1 was 0.979 in the UK Biobank test compared with 0.768 in the MrOS test. Likewise, the proportion of participants with accuracy (uncertainty) sufficient to evaluate hip fracture risk in Stage 1 was higher in the UK Biobank test (0.602) than in the MrOS test (0.532). For Stage 2, the UK Biobank test also showed higher accuracy for the UK Biobank than the MrOS data set (0.923 vs. 0.626), though the percentage of participants entering Stage 2 was slightly lower for UK Biobank than for MrOS (0.398 vs. 0.468).

For the female model (Trained on SOF), a similar trend for stage-wise accuracy was observed for Stage 1. The UK Biobank test (Table 7) showed higher accuracy for Stage 1 (0.839) compared to the SOF test (0.632), and a greater proportion of participants entered Stage 1 (0.531 vs. 0.477). For Stage 2, accuracy was slightly lower in the UK Biobank test (0.478) than in the SOF test (0.577), with a difference of 0.099. However, the percentage of participants entering Stage 2 was similar between the two datasets (0.469 for UK Biobank vs. 0.523 for SOF).

Overall, these results indicate that the two-stage model generalizes well to the UK Biobank external dataset, particularly for Stage 1, where both male and female models achieved consistently higher accuracy and participation rates compared to their respective internal tests (MrOS and SOF). In Stage 2, the UK Biobank performance still demonstrated favourable results in terms of accuracy and sample entry rates. For the female model, although the improvement in Stage 2 accuracy was modest, the UK Biobank results were generally consistent with those from the internal SOF test, further supporting the model's robustness and its ability to generalize across independent cohorts.

### 3.3.2 Two-Stage Model Demonstrates Reliable Performance Across Cohorts

Table 8 summarizes the performance metrics of the male and female DXA T-value two-stage models, trained on MrOS and SOF datasets, respectively, and evaluated on both internal and external (UK Biobank) test datasets. Key metrics—AUC, accuracy, sensitivity, and specificity—are reported with their means and standard deviations. For the male model (trained on MrOS), the UK Biobank test showed higher accuracy (0.955 vs. 0.707) and specificity (0.969 vs. 0.708), though the AUC was slightly lower (0.659 vs. 0.758). For the female model (trained on SOF), the UK Biobank test had a slightly higher AUC (0.659 vs. 0.652), along with improved accuracy (0.672 vs. 0.608) and specificity (0.678 vs. 0.605). For both models, the UK Biobank external test generally outperformed the internal test datasets in terms of accuracy, sensitivity, and specificity.

Further evaluation using 1,269 UK Biobank participants, detailed in Supplementary Tables 1–3 and 17, reinforces the two-stage model's strong external generalizability. In Tables 1-8, BMI was used as a predictor, while in Supplementary Tables 4–9, BMI was replaced with weight to align with variables used in the FRAX tool ([https://frax.shef.ac.uk/FRAX/](https://frax.shef.ac.uk/FRAX/)). Notably, BMI and weight are highly correlated (r = 0.91 for females, 0.87 for males), and the model's performance remained consistent regardless of which variable was used. Finally, despite differences in raw data with DXA device types - MrOS (Hologic 4500), SOF (Hologic 1000), and UK Biobank (GE-Lunar) - the two-stage model maintained stable performance across internal and external tests, as shown in Supplementary Tables 10–15. These findings collectively highlight the two-stage model's robustness and its ability to generalize effectively across diverse populations and imaging platforms.

# 4. Discussion

## 4.1 Two-Stage Model Reduces DXA Use While Preserving Predictive Accuracy Across Cohorts

To reduce clinical costs by minimizing unnecessary DXA imaging, we developed a sequential two-stage model that combines predictions from Ensemble 1 (Stage I) and Ensemble 2 (Stage II) to predict hip fracture risk. Ensemble 1 uses only clinical variables, while Ensemble 2 incorporates both clinical and DXA imaging data. If an individual is predicted to have low or high risk for hip fracture with high confidence in Stage I, he/she would not need to obtain DXA imaging. Otherwise, the individual will be referred for DXA imaging which would be incorporated into the Stage 2 model to evaluate his/her risk for hip fracture.

Our results from both internal and external validation datasets showed that the two-stage model achieved higher AUC, sensitivity, and specificity than the model using only clinical variables or the model combining clinical variables with DXA imaging alone. This improvement arises because the uncertainty rule in our advanced two-stage framework retains individuals with high prediction certainty in Stage 1, while directing those with low certainty to Stage 2 for further assessment—thereby enhancing overall sensitivity. Clinical variables alone cannot provide sufficient certainty for participant classification due to the limited information they contain. Even if the two-stage model shows slightly lower accuracy than the model combining clinical variables and DXA imaging, it offers substantial advantages by reducing imaging costs and identifying individuals who truly require imaging at an earlier stage.. In addition, our results showed that only a small percentage of subjects were  included in Stage 2, so only this small subset of individuals would require DXA imaging. Therefore, our two-stage model provide a practical strategy to reduce the clinical cost by only performing the DXA imaging for participants in Stage II with little reduction in accuracy for hip fracture risk assessment.

External validation using the UK Biobank dataset confirms our two-stage model's generalizability beyond its training cohorts (MrOS and SOF). This staged model maintained strong predictive performance across varied populations and settings, demonstrating its adaptability and reliability. These findings support the model's clinical utility and highlight its ability to provide accurate fracture risk assessment across diverse cohorts.

### 4.2 Two-Stage Model Outperforms Traditional T-Score Classification in Sensitivity

We compared the performance of our two-stage model with the traditional clinical classification of osteoporosis based on Total BMD T-scores. Because T-scores are not available in the MrOS and SOF datasets, we relied on T-values for comparative reference. According to the World Health Organization (WHO), a T-score $\geq -1$ indicates normal bone density, $-1$ to $-2.5$ indicates osteopenia, and $\leq -2.5$ indicates osteoporosis[21] (Cosman et al., 2014). Using these thresholds, the sensitivity of T-score-based osteoporosis prediction was low: for the MrOS dataset, sensitivity was 0.45 with a cutoff of $-1$ and only 0.025 with a cutoff of $-2.5$; for the SOF dataset, sensitivity was 0.279 (cutoff $-1$) and 0.011 (cutoff $-2.5$). In contrast, our two-stage model achieved much higher sensitivity - 0.717 for MrOS and 0.678 for SOF - demonstrating a significant improvement over traditional T-score-based diagnosis for osteoporosis and the subsequent hip fracture risk prediction. Given the relatively low prevalence of hip fractures in the general population, early detection of true positives is vital for timely intervention and prevention.

**4.3 Two-Stage Model Achieves Higher Sensitivity and Competitive Accuracy Compared to FRAX**

We also compared the performance of our two-stage model with FRAX (T-score BMD-based) predictions using three cleaned datasets: MrOS (Hologic 4500), SOF (Hologic 1000), and UK (GE-Lunar). Each dataset includes key FRAX variables - age, weight, height, femoral neck BMD, smoking status, and prior fractures - in the required order for direct comparison. FRAX showed lower sensitivity across all datasets: 0.102 for MrOS (Hologic 4500), 0.245 for SOF (Hologic 1000), and 0.051 for UK (GE-Lunar). In contrast, our two-stage model achieved notably higher sensitivity: 0.682 for MrOS, 0.637 for SOF, and 0.072 for UK. Specifically for the UK male group, the two-stage model outperforms FRAX in all key metrics—AUC (0.644 vs. 0.596), accuracy (0.979 vs. 0.897), sensitivity (0.072 vs. 0.051), and specificity (0.992 vs. 0.989). In addition to its superior sensitivity, the two-stage model demonstrated comparable performance to FRAX across other key metrics for MrOS, SOF and the UK female group, with only minor differences. Because hip fractures are relatively uncommon in the general population, accurately detecting true positives early is essential for prompt treatment and prevention. Our method's high sensitivity underscores its effectiveness in identifying at-risk individuals early, enabling timely and targeted interventions.

**4.4 Two-Stage Model Reduces Costs, Preserves Interpretability, and Enables Future Enhancements**

 Our two-stage approach offers substantial benefits in reducing healthcare costs and streamlining screening for hip fracture risk. Conventional methods for fracture risk assessment typically require DXA imaging for most individuals, which increases financial burden and is inconvenient. In contrast, our two-stage model begins with clinical variables to identify high-confidence cases, limiting the need for imaging to only those who require further evaluation. This targeted use of DXA in the second stage significantly decreases unnecessary DXA scans and associated costs.

Another notable strength of our model is its interpretability - it provides clear understanding of the key predictors associated with hip fracture risk. Many of these predictors, such as age, weight, height, grip strength, BMI, walking speed, mental status, daily functioning, smoking behavior, and DXA-derived features, align with established risk factors documented in the literature. By integrating this broad range of variables, the model captures the complex nature of fracture risk and supports a more holistic assessment. This approach enhances early identification of high-risk individuals, enabling timely, personalized preventive strategies that can reduce the incidence of fractures and improve clinical outcomes.

Additionally, the staged design lends itself to future enhancements. Its flexible structure allows for the incorporation of new data sources—such as genetic data[22] (Nethander et al., 2022) or advanced biomarker data - offering opportunities to improve predictive accuracy and individualize risk profiles even further. As such, our model not only embodies current best practices in fracture risk prediction but also establishes a scalable and adaptive framework, ready to evolve alongside advances in medical research and data availability. This adaptability strengthens its long-term utility in real-world clinical settings.

## 5. Conclusion

This study presents a sequential two-stage model for hip fracture risk prediction that improves upon traditional approaches such as T-score and FRAX. By first identifying individuals unlikely to be at risk and then applying a refined model to the remaining high-risk population, our framework achieves higher sensitivity while simultaneously reducing unnecessary DXA screening. Internal and external validation across multiple independent cohorts confirmed the robustness, accuracy, and adaptability of this approach, demonstrating its potential to be applied broadly in diverse clinical and research settings. Observing consistent results across the two measures of harmonized BMD data and T-value BMD further support the reliability of our approach and establish a comprehensive framework for assessing the internal and external validity of the two-stage model strategy. Importantly, the staged design offers straightforward interpretability and practical clinical utility, which are essential for adoption in real-world health systems. In addition to immediate applications in hip fracture risk assessment and fracture prevention, this framework is flexible and can be extended to incorporate genetic and biomarker data, providing a foundation for even more precise risk stratification. Overall, this work highlights a scalable, cost-effective strategy with the potential to improve outcomes and reduce the burden of osteoporosis and fractures in aging populations.

## Data Availability

The data used in this study cannot be shared with the public due to third-party use restrictions and patient confidentiality concerns. The three datasets utilized in this study were obtained from dbGaP and UK Biobank websites, the links are:

The Osteoporotic Fractures in Men (MrOS) (phs000373.v1.p1)

https://www.ncbi.nlm.nih.gov/projects/gap/cgi-bin/study.cgi?study_id=phs000373.v1.p1

The Study of Osteoporotic Fractures (SOF) (phs000510.v1.p1)

https://www.ncbi.nlm.nih.gov/projects/gap/cgi-bin/study.cgi?study_id=phs000510.v1.p1

The UK Biobank (application ID: 61915)

https://www.ukbiobank.ac.uk/enable-your-research/about-our-data

MrOS and SOF represent the male and female groups, respectively, with data obtained under the Disease-Specific (Aging Related 1, RD) consent group.

## Code Availability

All statistical analyses were conducted using R software (version 4.3.0; R Foundation for Statistical Computing, Vienna, Austria), The code used for this study is available at: https://github.com/MIILab-MTU/StagedModelForHipFractureRWithMultiDatasetsiskPrediction

## Author Information

These authors contributed equally: Shuo Sun, Meiling Zhou

These authors jointly supervised this work: Kui Zhang, Weihua Zhou.

Affiliations

**Department of Mathematical Sciences, Michigan Technological University, Houghton, MI, USA**

Shuo Sun, Kui Zhang*

**Department of Statistics, Grand Valley State University, Allendale, MI, USA**

Meiling Zhou

**Department of Computer Science, Kennesaw State University, Marietta, GA, USA**

Chen Zhao

**Department of Radiological Sciences, Department of Biomedical Engineering, and Department of Mechanical and Aerospace Engineering, University of California, Irvine, CA, USA**

Joyce H. Keyak

**Department of Internal Medicine and Division of Rheumatology, UC Davis Health, Sacramento, CA**

Nancy E. Lane

**Geisel School of Medicine, Dartmouth College, Hanover, NH, USA**

Jeffrey D. Deng

**Division of Biomedical Informatics and Genomics, Tulane Center of Biomedical Informatics and Genomics, Deming Department of Medicine, Tulane University, New Orleans, LA 70112**

Kuan-Jui Su, Hui Shen & Hong-Wen Deng

**Department of Applied Computing, Michigan Technological University, Houghton, MI, USA**

Weihua Zhou[*]

Shuo Sun[1], Meiling Zhou[2], Chen Zhao[3], Joyce H. Keyak[4], Nancy E. Lane[5], Jeffrey D. Deng[6], Kuan-Jui Su[7], Hui Shen[7], Hong-Wen Deng[7], Kui Zhang[1,*], Weihua Zhou[8,9*]

## Contributions

**Shuo Sun** contributed to Writing – review & editing, Validation, Software, Methodology, and Conceptualization. **Meiling Zhou** contributed to Writing – review & editing, Validation, Software, Methodology, Conceptualization and Formal analysis. **Chen Zhao** contributed to review & editing and conducting FRAX website predictions across the three datasets. **Joyce H. Keyak** contributed to Writing - review & editing, Methodology, and Formal analysis. **Nancy E. Lane** contributed to Writing - review & editing, validation, and Conceptualization. **Jeffrey D. Deng** contributed to Writing - review & editing. **Kuan-Jui Su** contributed to Writing - review & editing and data access and project management. **Hui Shen** contributed to Project administration. **Hong-Wen Deng** contributed to Writing - review & editing, and conceptualization. **Kui Zhang** contributed to Writing – review & editing, Supervision, Project administration, Software, Methodology, Formal analysis, and Conceptualization. **Weihua Zhou** contributed to Writing – review & editing, Validation, Supervision, Project administration, Methodology, Formal analysis, and Conceptualization.

## Corresponding authors

Correspondence to:  Kui Zhang and Weihua Zhou.

ORCID IDs:

Kui Zhang: [0000-0002-2441-2064]

Weihua Zhou: [0000-0002-6039-959X]

## Acknowledgements

This research has been conducted using the UK Biobank Resource under application number [61915]. It was in part supported by grants from the National Institutes of Health, USA (U19AG055373, 1R15HL172198, and 1R15HL173852) and American Heart Association (#25AIREA1377168).

## Ethics declarations

### Competing interests

All authors declare that there are no conflicts of interest.

## Supplementary Information

Supplementary Information

# Tables

**Table 1.** Distribution of Hip Fractures by Racial Group in the MrOS Male Cohort and the SOF Female Cohort.

| Male (MrOS) | No Fracture | Fracture |
|---|---|---|
| White | 3699 (96.1%) | 150 (3.9%) |
| African American | 118 (97.5%) | 3 (2.5%) |
| Asian | 164 (99.4) | 1 (0.6%) |
| Hispanic | 107 (96.4%) | 4 (3.6%) |
| Other | 46 (93.9%) | 3 (6.1%) |
| **Female (SOF)** | **No Fracture** | **Fracture** |
| White | 3045 (84.6%) | 553 (15.4%) |
| Asian | 4 (80.0%) | 1 (20.0%) |
| Hispanic | 2 (66.7%) | 1 (33.3%) |
| Other | 2 (1.0%) | 0 (0.0%) |

**Table 2.** Hip Fracture Distribution in UK Biobank External Test Datasets.

| | Age: 54-70 | male | female |
|---|---|---|---|
| UK Biobank external test dataset 1 (1269 participants) | No hip fracture | 396 | 861 |
| | Hip fracture | 6 | 6 |
| UK Biobank external test dataset 2 (625 participants) | No hip fracture | 208 | 409 |
| | Hip fracture | 3 | 5 |

**Table 3.** Performance of the two-stage model using harmonized BMD with MrOS Male Data.

| Model | | AUC | Accuracy | Sensitivity | Specificity |
|---|---|---|---|---|---|
| Ensemble 1 | Mean (SD) | 0.699 (0.002) | 0.649 (0.021) | 0.340 (0.023) | 0.660 (0.023) |
| Two-Stage | Mean (SD) | **0.807** (0.039) | **0.759** (0.051) | **0.712** (0.120) | **0.761** (0.055) |
| Ensemble 2 | Mean (SD) | 0.829 (0.002) | 0.705 (0.019) | 0.279 (0.021) | 0.721 (0.021) |
| Accuracy for participants kept in stage 1 | | | 0.846 | | |
| Percentages of participants kept in stage 1 | | | 0.615 | | |
| Accuracy for participants kept in stage 2 | | | 0.608 | | |
| Percentages of participants kept in stage 2 | | | 0.385 | | |

**Table 4.** Performance of Two-Stage Model using T-value BMD with MrOS Male Data.

| MrOS Male | Model | AUC | Accuracy | Sensitivity | Specificity |
|---|---|---|---|---|---|
| Ensemble 1 | Mean (SD) | 0.699 (0.002) | 0.648 (0.023) | 0.341 (0.025) | 0.659 (0.025) |
| Two-Stage | Mean (SD) | **0.806** (0.040) | **0.759** (0.054) | **0.717** (0.106) | **0.760** (0.058) |
| Ensemble 2 | Mean (SD) | 0.829 (0.002) | 0.707 (0.020) | 0.277 (0.021) | 0.723 (0.021) |
| Accuracy for keeping in stage 1 | | | 0.846 | | |
| Percentages of participants keeping in stage 1 | | | 0.623 | | |
| Accuracy for keeping in stage 2 | | | 0.606 | | |
| Percentages of participants keeping in stage 2 | | | 0.377 | | |

**Table 5.** Performance of two-stage model using harmonized BMD with SOF Female Data.

| SOF Female | Model | AUC | Accuracy | Sensitivity | Specificity |
|---|---|---|---|---|---|
| Ensemble 1 | Mean (SD) | 0.612 (0.002) | 0.512 (0.018) | 0.483 (0.025) | 0.517 (0.025) |
| Two-Stage | Mean (SD) | **0.761** (0.023) | **0.682** (0.068) | **0.698** (0.112) | **0.679** (0.098) |
| Ensemble 2 | Mean (SD) | 0.776 (0.001) | 0.593 (0.015) | 0.366 (0.021) | 0.634 (0.021) |
| Accuracy for keeping in stage 1 | | | 0.727 | | |
| Percentages of participants keeping in stage 1 | | | 0.745 | | |
| Accuracy for keeping in stage 2 | | | 0.545 | | |
| Percentages of participants keeping in stage 2 | | | 0.255 | | |

**Table 6.** Performance of Two-Stage Model using T-value BMD with SOF Female Data.

| SOF Female | Model | AUC | Accuracy | Sensitivity | Specificity |
|---|---|---|---|---|---|
| Ensemble 1 | Mean (SD) | 0.613 (0.002) | 0.511 (0.016) | 0.484 (0.023) | 0.516 (0.023) |
| Two-Stage | Mean (SD) | **0.761** (0.021) | **0.697** (0.066) | **0.678** (0.118) | **0.701** (0.097) |
| Ensemble 2 | Mean (SD) | 0.779 (0.001) | 0.593 (0.016) | 0.366 (0.023) | 0.634 (0.023) |
| Accuracy for keeping in stage 1 | | | 0.743 | | |
| Percentages of participants keeping in stage 1 | | | 0.752 | | |
| Accuracy for keeping in stage 2 | | | 0.554 | | |
| Percentages of participants keeping in stage 2 | | | 0.248 | | |

**Table 7.** Internal vs. External (UK Biobank 625 participants) Two-Stage Model Test Results for Male and Female T-value DXA (no hip fracture follow-up time).

| Two-Stage Model | Average accuracy | | Average percentage | |
|---|---|---|---|---|
| **Train: MrOS** | **Test UK biobank** | **Test MrOS** | **Test UK biobank** | **Test MrOS** |
| Enter into Stage 1 | 0.979 | 0.768 | 0.602 | 0.532 |
| Enter into Stage 2 | 0.923 | 0.626 | 0.398 | 0.468 |
| **Train: SOF** | **Test UK biobank** | **Test SOF** | **Test UK biobank** | **Test SOF** |
| Enter into Stage 1 | 0.839 | 0.632 | 0.531 | 0.477 |
| Enter into Stage 2 | 0.478 | 0.577 | 0.469 | 0.523 |

**Table 8.** Performance Metrics for Male and Female DXA T-value Two-Stage Models (MrOS and SOF Training Sets) Across Internal and External (UK Biobank) Test Datasets (no hip fracture follow-up time).

| Metric | Mean | | SD | |
|---|---|---|---|---|
| **Train: MrOS** | **Test UK biobank** | **Test MrOS** | **Test UK biobank** | **Test MrOS** |
| AUC | 0.659 | 0.758 | 0.075 | 0.042 |
| Accuracy | 0.955 | 0.707 | 0.021 | 0.076 |
| Sensitivity | 0.000 | 0.688 | 0.000 | 0.134 |
| Specificity | 0.969 | 0.708 | 0.021 | 0.083 |
| **Train: SOF** | **Test UK biobank** | **Test SOF** | **Test UK biobank** | **Test SOF** |
| AUC | 0.659 | 0.652 | 0.061 | 0.028 |
| Accuracy | 0.672 | 0.608 | 0.084 | 0.061 |
| Sensitivity | 0.174 | 0.625 | 0.073 | 0.110 |
| Specificity | 0.678 | 0.605 | 0.085 | 0.090 |

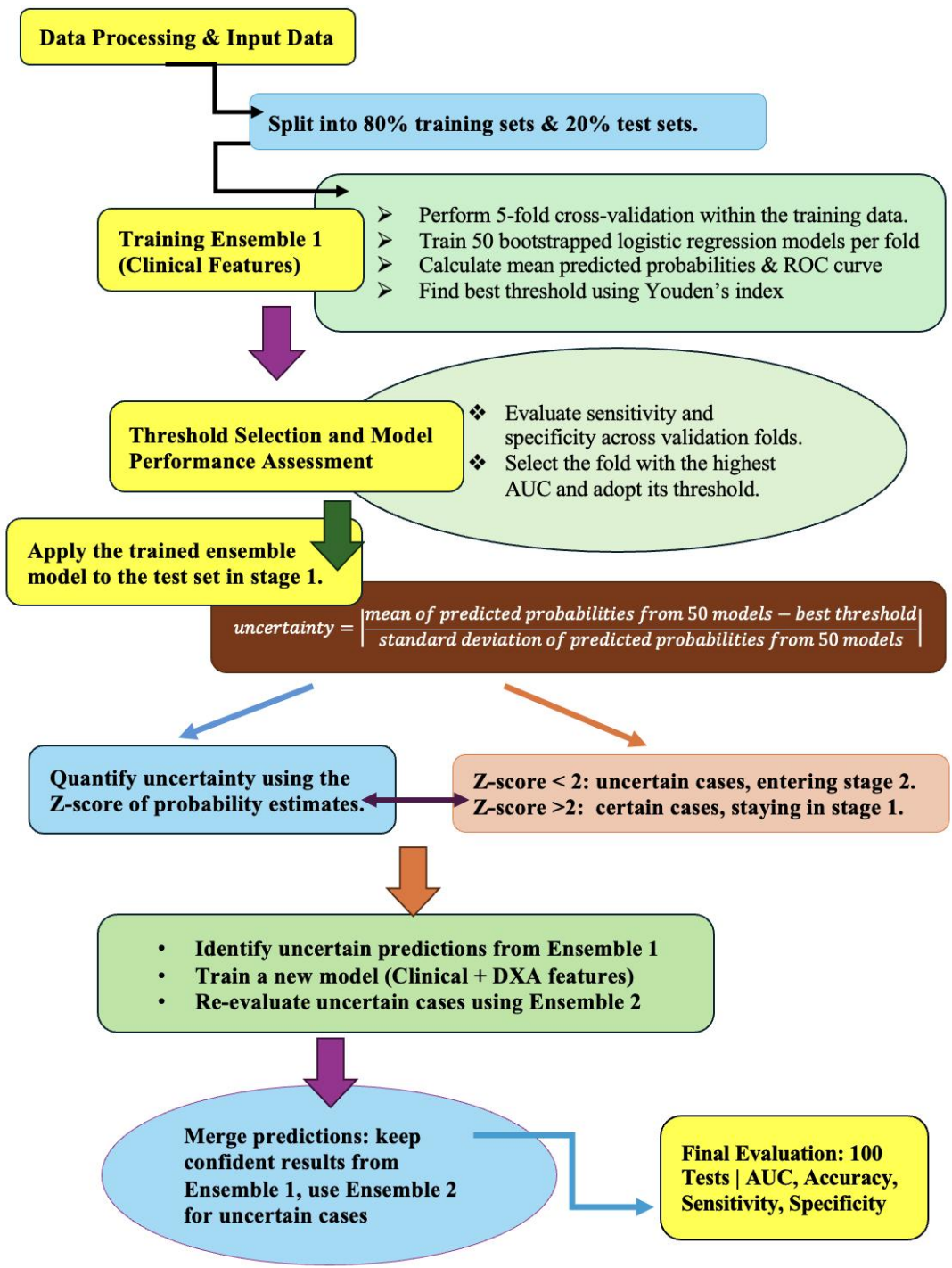

**Figure 1.** Flowchart of Advanced Two-Stage Model Internal & External Test

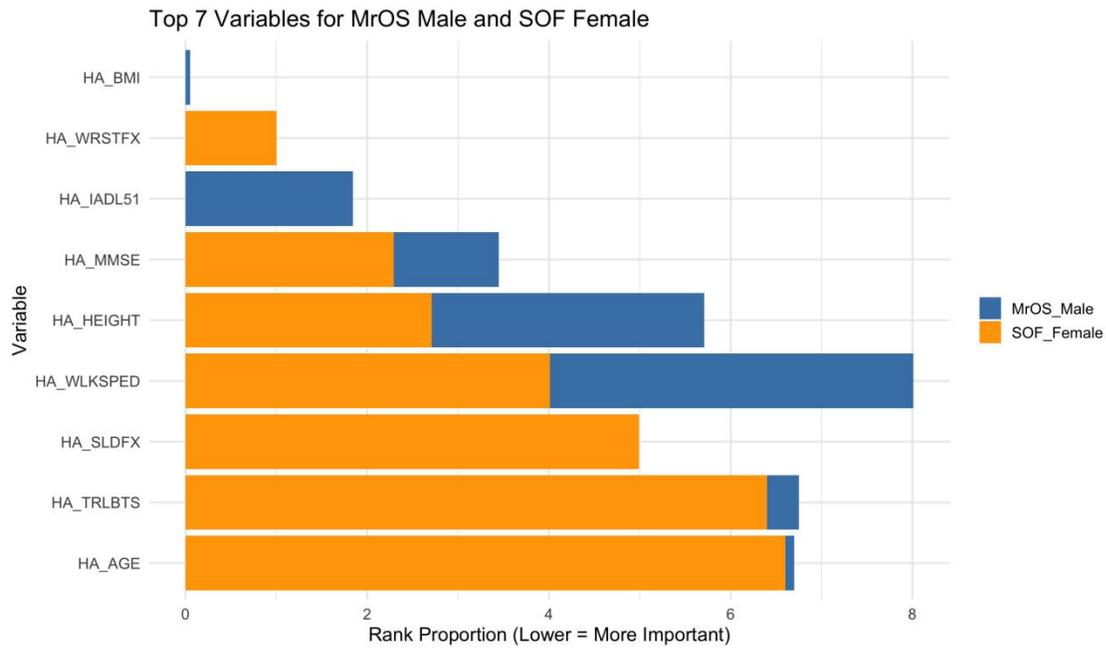

**Figure 2.** Top 7 variables for MrOS Male and SOF Female in Stage 1. Lowest Rank Proportion (most important) at the top of the plot. See Supplementary Tables for more information about the variables.

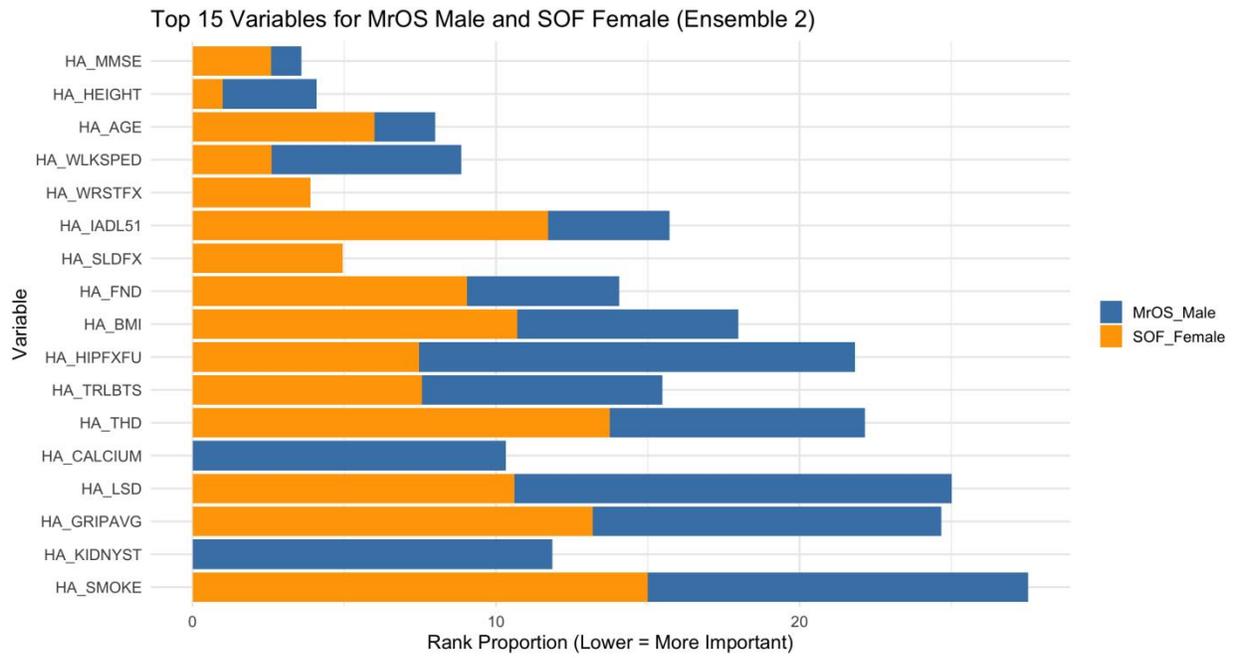

**Figure 3.** Top 15 variables for MrOS Male and SOF Female in Stage 2. Lowest Rank Proportion (most important) at the top of the plot. See Supplementary Tables for more information about the variables.